\documentclass[letterpaper, 10 pt, conference]{ieeeconf}  % Comment this line out if you need a4paper

\IEEEoverridecommandlockouts                              % This command is only needed if 
\usepackage{graphicx}
\usepackage{array}
\usepackage{ wasysym }
\usepackage{amssymb} %for checkmark
\usepackage{booktabs, threeparttable}

\usepackage{tikz}
\usepackage{textcomp}
\usepackage{hyperref}
\usepackage{lipsum}

\newcommand\copyrighttext{%
  \footnotesize \textcopyright \ 2022 IEEE.  Personal use of this material is permitted.  Permission from IEEE must be obtained for all other uses, in any current or future media, including reprinting/republishing this material for advertising or promotional purposes, creating new collective works, for resale or redistribution to servers or lists, or reuse of any copyrighted component of this work in other works.}
\newcommand\copyrightnotice{%
\begin{tikzpicture}[remember picture,overlay]
\node[anchor=south,yshift=10pt] at (current page.south) {\fbox{\parbox{\dimexpr\textwidth-\fboxsep-\fboxrule\relax}{\copyrighttext}}};
\end{tikzpicture}%
}

\title{\LARGE \bf
MetaGraspNet: A Large-Scale Benchmark Dataset for Scene-Aware Ambidextrous Bin Picking via Physics-based Metaverse Synthesis 
}
\author{Maximilian Gilles*$^{1}$, Yuhao Chen*$^{2}$, Tim Robin Winter$^{1}$, E. Zhixuan Zeng$^{2}$, Alexander Wong$^{2}$
\thanks{*Both authors contributed equally.}% <-this % stops a space
\thanks{$^{1}$Karlsruhe Institute of Technology (KIT), Karlsruhe, Germany
        {\tt\small maximilian.gilles@kit.de}}%
\thanks{$^{2}$University of Waterloo (UW), Waterloo,
        Canada
        {\tt\small yuhao.chen1@uwaterloo.ca}}
\thanks{This research has been founded by German Federal Ministry for Economic Affairs and Climate Action (BMWK) and the National Research Council Canada (NRC).}
}
\begin{document}
\maketitle
\copyrightnotice
\thispagestyle{empty}
\pagestyle{empty}

%%%%%%%%%%%%%%%%%%%%%%%%%%%%%%%%%%%%%%%%%%%%%%%%%%%%%%%%%%%%%%%%%%%%%%%%%%%%%%%%
\begin{abstract}
Autonomous bin picking poses significant challenges to vision-driven robotic systems given 
%An increasingly important use-case for robotic systems in modern smart warehouses and factories is autonomous bin picking. This task poses significant challenges given
the complexity of the problem, ranging from various sensor modalities, to highly entangled object layouts, to diverse item properties and gripper types. 
%Existing bin picking solutions often address the problem from one perspective such as finding the best grasping point without considering how objects stack to prevent damage.
Existing methods % are versatile but 
 often address the problem from one perspective. 
%but do consider the specific requirements bin picking poses such as the usage of vacuum technology or the strong interaction of objects with the bon.
%without dealing with bin picking specific problems.
Diverse items and complex bin scenes require diverse picking strategies together with advanced reasoning.
%For example, %parallel gripper is better for picking pens,
%vacuum gripper is better than parallel gripper for picking boxes. %, and RGB sensor is better for detecting transparent objects. 
%such as ambidextroux picking  order  
%without considering the 
 As such, to build robust and effective machine-learning algorithms for solving this complex %robotic grasp problem
task requires significant amounts of comprehensive and high quality data. Collecting such data in real world would be too expensive and time prohibitive and therefore intractable from a scalability perspective. 
To tackle this big, diverse data problem, we take inspiration from the recent rise in the concept of metaverses, and introduce MetaGraspNet, a large-scale photo-realistic bin picking dataset constructed via physics-based metaverse synthesis. The proposed dataset contains 217k RGBD images across 82 different article types% and 37 viewpoints
, with full annotations for object detection, amodal perception, keypoint detection, manipulation order and ambidextrous grasp labels for a parallel-jaw and vacuum gripper. We also provide a real dataset consisting of over 2.3k fully annotated high-quality RGBD images,
%captured with an industry-grade camera system.
divided into 5 levels of difficulties and an unseen object set to evaluate different object and layout properties. Finally, we conduct extensive experiments showing that our proposed vacuum seal model and synthetic dataset achieves state-of-the-art performance and generalizes to real world use-cases.
\end{abstract}
%%%%%%%%%%%%%%%%%%%%%%%%%%%%%%%%%%%%%%%%%%%%%%%%%%%%%%%%%%%%%%%%%%%%%%%%%%%%%%%%
% scope, novelty, significance %
\section{INTRODUCTION}
\label{sec:intro}
Bin picking with its central role in automation and logistics is an important use-case for autonomous robotic systems in today's smart warehouses or factories. %It reduces labor from performing repetitive and sometimes physically demanding picking tasks.
Many existing commercial systems are able to pick and move objects autonomously by simplifying the task, carefully restricting the item set or structuring the grasp environments. % such as arranging items into fixed locations prior to picking. 
To advance bin picking into the next level, robotic systems must begin to understand the bin scene in order to handle more complex and diverse scenarios, dealing with items differ in shape, color, texture, pose, and dealing with object layouts differ in density and entanglement. % One example application would be sorting recycled items.
In such autonomous systems, vision plays an important role to identify items as well as their poses, grasping points and manipulation order. 
However, the high complexity of bin scenes as well as the wide range of different articles present great challenges which limit the applicability of today's robotic systems.

%Tasks of picking systems part of modern assembly or warehouse robots include targeted grasping of complex items in highly unstructured environments and their accurate placement.
%For the successful completion of this high-level task, the system has to be aware of the objects in the scene and understand the underlying scene layout in order to find reliable grasps and plan a collision-free path. 
%Typical use cases require the robot to empty the tote or to pick and place a specific item in clutter. Picking items from top to bottom is crucial for avoiding uncontrolled item movements that leads to damages. 
%In both cases the vision system has to be aware of the objects in the scene and understand the underlying scene layout in order to avoid false grasp attempts, collision with the scene or object dropping.

   \begin{figure}[t]
    \centering
    \includegraphics[width=\columnwidth]{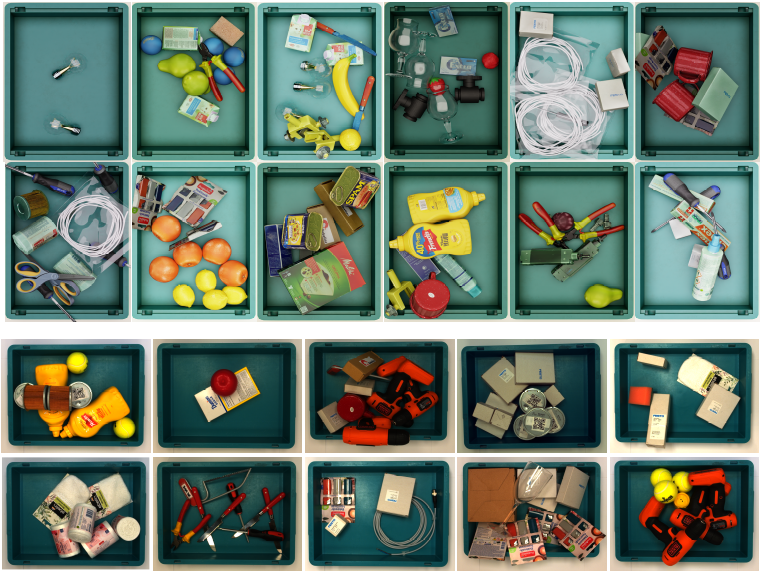}
    \caption{Synthetic scenes generated in metaverse (top 2 rows) and real world evaluation dataset (bottom 2 rows). Both datasets are captured from multiple camera viewpoints and provide fully comprehensive ground truth for parallel-jaw and vacuum gripper and scene reasoning.}
     \label{dataset_scenes}
     \vspace{-6mm}
   \end{figure}

In this work, we address the vision problem of bin picking in two parts: finding targeted objects, predicting reliable grasp points for the objects.
We formulate the object finding as an object detection problem and associate 3 challenges with it in the bin picking context.
%. There are 3 challenges associated with object detection in bin picking. 
The first challenge is combining information from multiple modality. Modern robotic grasping systems are equipped with multiple imaging sensors. The most popular ones are RGB and depth. RGB sensor captures the fine details of object's texture, while depth sensor captures object's surface location and thus providing excellent geometry information. However, each sensor has its own drawback. RGB sensor is susceptible to shadow, and objects with similar textures are difficult to differentiate. Depth sensor is prone to noise and produces faulty or invalid values for transparent and reflective objects. Leveraging the advantage of both sensor types is a non-trivial problem.
The second challenge is scene understanding, namely knowing where objects are and how they are posed and stacked. 
Objects in cluttered bin scenes are heavily occluded and entangled, which reduces the amount of information that can be used to estimate objects' pose and their stacking relationship.
In addition, it is very common to have multiple instances of the same class randomly stacking in a bin, resulting an object visually breaking into multiple parts, which is prone to being detected as multiple objects. 
The third challenge is unseen objects, or objects that changes shapes. To detect an unseen object requires the vision system to understand objects at a basic texture and geometry feature level.
When objects are scattered in the bin, visual features can be separated spatially. 
When multiple instances of unseen objects stacking together, visual features of the same class entangle at one location, and spatial information is not enough to separate instances.

We formulate grasp points finding as a grasp detection problem. 
Commercial gripper types includes parallel gripper, suction gripper, variations of both or even 5-finger gripping hands. Each gripper type deals with items with specific shape, and texture. 
For example, suction gripper can pick box items and bags well, but struggles with complex or filigree objects. 
With a high diversity of items, there is no single gripper type that is suitable for picking all the items. Highly flexible robotic systems can have multiple gripper types and choosing the right one is non-trivial. 
%The high diversity of layout and item properties, as well as the complex physical interaction between gripper and item poses great challenges on the grasp detection. 
A successful grasp is highly dependent on the object's properties such as local surface property and structure, mass distribution as well as the object's relationship with other objects in the bin. Finding reliable grasps is therefore not limited to 
%the physical properties of each type of gripper
reason about the physical interaction between gripper and object, but also challenges the system to understand the arrangement of objects in the scene and identify an appropriate picking sequence.
%The subsequent precise placement of a picked item goes beyond that, the system has to be aware of how the item was grasped beforehand in order to place it accurately. 
%For unstructured environments this can be formulated as pose estimation problem.
While many researches have provided excellent datasets on each individual part such as SynPick \cite{SynPick} for pose estimation and gripper-object interaction, REGRAD \cite{Zhang.29.04.2021} for relationship reasoning, SuctionNet-1Billion \cite{Cao.2021} for vacuum grasping or Dex-Net 4.0 \cite{Mahler.2019} for ambidextrous grasping, the problems were only addressed from one side or did not target automation. Because robotic picking is a multi-stage complex problem, solving the problem from one aspect will miss the details of other aspects. Therefore, it is important to have a comprehensive dataset that covers all the aspects of robotic bin picking.

Collecting such comprehensive data from real robot experiments \cite{Kalashnikov.27.06.2018, Levine.2018} or manually \cite{Zeng.2019} would be too expensive and time prohibitive and as such intractable from a scalability perspective. Motivated by earlier work in the field of synthetic data generation \cite{SynPick, Zhang.29.04.2021} and inspired from the recent rise in the concept of metaverses, we introduce MetaGraspNet: a large-scale photo-realistic physics-based bin picking dataset for ambidextrous grasping and bin scene reasoning.
%We propose a complete synthetic data generation pipeline for robot bin picking with a vacuum gripper and parallel gripper.  In contrast to many other work \cite{Zhang.2021, Mahler.2019, C.Eppner.2020} our proposed dataset consists of  photo-realistic cluttered scenes together with rich scene annotations suitable for ambidextrous grasping as well as scene reasoning. 
By providing rich semantic scene labels such as amodal segmentation masks or object layout graphs together with heterogeneous grasp labels, object poses, and keypoint labels, MetaGraspNet challenges picking systems to take the next step towards multi-gripper usage and cognitive understanding of bin scenes.

Our contributions can be summarized as following:
\begin{itemize}

\item We contribute 36 new objects
%high-quality 3D scans of objects 
(including transparent and specular items) suitable for a vacuum or parallel gripper.
\item We propose a force-based suction cup model able to predict the vacuum seal for grasp candidates  and provide a thorough method for generating parallel-jaw grasps based on physics simulation. 
\item We contribute a pipeline to generate photo-realistic bin picking scenes together with a large-scale dataset. Besides rich grasp label annotation, it provides segmentation mask, object pose, center of mass heatmaps and propose the concept of semantic object keypoints.
\item We present labels to characterize the objects' layout in the bin: amodal segmentation masks, occlusion rates, object relationship matrix and layout label. 
\item For evaluation we provide a real dataset consisting of 2.3k pixel-wise annotated RGBD images captured in a logistic setting with an industry-grade camera system.
\item We conduct extensive experiments in  real world evaluating our proposed vacuum seal model and providing baseline experiments for vacuum grasp point detection and object detection and segmentation.
\end{itemize}

Our synthetic and real dataset, as well as the complete codebase to generate custom data are publicly available at {\footnotesize \fontfamily{qcr}\selectfont
https://github.com/maximiliangilles/MetaGraspNet}.

\begin{table*}[tbp]
\caption{Overview over existing RGBD grasp datasets}
\vspace{-6mm}
\begin{threeparttable}
\begin{center}
\begin{tabular*}{\textwidth}{@{\extracolsep{\fill}}l cccc cccccccc}
\toprule
Work&\multicolumn{3}{c}{Data}& &\multicolumn{8}{c}{Grasp and Scene Labels\tnote{a}} \\
\cmidrule{2-4}
\cmidrule{6-13}
 & Item Sets  & Sensor Type & Scene Comp. & 
 & DoF  & PJ\tnote{b}  & SC\tnote{b} & Pose  & Amod.
  & Layout  & CoM  & Keypts\\
\midrule
\cite{YunJiang.2011} & custom & real & single object & & 4 & hand &  &  &  &  &  &    \\
\cite{A.Depierre.2010} & \cite{shapenet2015} & sim & single object & & 6 & sim &  &  &  &  &  &    \\
\cite{H.Zhang.2019} & custom & real & cluttered & & 4 & hand &  &  &  & \checkmark &  &    \\
\cite{Fang.62020} & custom + \cite{B.Calli.2015} & real & cluttered & & 6 & analytic &  & \checkmark &  &  &  &    \\
\cite{Cao.2021} & custom + \cite{B.Calli.2015} & real & cluttered & & 6 &  & analytic & \checkmark &  &  &  &    \\
\cite{Zhang.29.04.2021} & \cite{shapenet2015} & sim + real & highly cluttered & & 6 & analytic &  & \checkmark &  & \checkmark &  &    \\
\cite{SynPick} & \cite{xiang2018posecnn} & sim & highly cluttered & & 6 &  & segmap & \checkmark &  & \checkmark &  &    \\
\cite{Zeng.2019} & custom & real & highly cluttered & & 4 & hand & hand &  &  &  &  &    \\
\addlinespace
\textbf{ours} & \textbf{custom + \cite{B.Calli.2015} + \cite{Liu.23.04.2021}} & \textbf{sim + real} & \textbf{ highly cluttered} &  & \textbf{6} & \textbf{sim} & \textbf{analytic} & \textbf{\checkmark} & \textbf{\checkmark} & \textbf{\checkmark} & \textbf{\checkmark} & \textbf{\checkmark}   \\
\bottomrule

%\multicolumn{12}{l}{\textit{PJ}: parallel-jaw grasp label aggregation, \textit{SC}: suction cup grasp label aggregation, \textit{Amod. Perc.}: amodal perception, \textit{CoM}: center of mass heatmap}\\
%\multicolumn{12}{l}{\textit{hand}: manually labeled by hand, \textit{sim}: physics simulation, \textit{analytic}: analytical sampling strategies , \textit{segmap}: 2d segmentation mask and heuristic}\\

%\multicolumn{4}{l}{$^{\mathrm{*}}$: medium cluttered, $^{\mathrm{**}}$: highly cluttered}
\end{tabular*}
\label{table_overview_datasets}
\scriptsize
\begin{tablenotes}
%\RaggedRight
\item[a] DoF: Degree of Freedom Grasp;
         PJ: Parallel-Jaw Label Aggregation; 
         SC: Vacuum Grasp Label Aggregation; 
         Amod.: Amodal Segmentation Mask;
         CoM: Center of Mass Heatmap;
         Keypts.: Keypoint Label
\item[b] hand: manually labelled;
         sim: physics simulation;
         analytic: analytical sampling;
         segmap: 2D segmentation mask and heuristic
\end{tablenotes}
\vspace{-6mm}
\end{center}
\end{threeparttable}
\end{table*}

\section{Related Work}

Datasets for robotic grasping are versatile and differ in many aspects such as scene composition, item diversity, sensor modality, gripper types or labelled properties. Table~\ref{table_overview_datasets} can be seen as an attempt to give a general overview over existing RGBD datasets and its labelled properties.
%by updating and fusing already existing work \cite{C.Eppner.2020, Fang.62020, Cao.2021}
%usually restricted to parallel-jaw gripper (PJ) \cite{C.Eppner.2020,Fang.62020} or suction cup (SC) \cite{Cao.2021} grippers%
%and enriching them with additional taxonomy useful for scene-aware bin picking.
%such as scene composition, semantic keypoints, object poses, occlusion or center of mass heatmap.

%\subsection{Object Sets and Photorealism} \label{objects_set}
%Robotic grasping datasets differ in sensor modality and in object datasets used for generating the scenes. 

\medskip
\noindent
\textbf{Object Sets and Photorealism:} Motivated by recent progress in the field of dexterous grasping based on RGBD data \cite{Gou.03.03.2021}, we find it beneficial to categorize existing work into depth only or photo-realistic data. While depth only datasets \cite{Zhang.2021, Mahler.2019,C.Eppner.2020,Mahler.20.09.2017,Kleeberger.12.01.2021,Morrison.03.03.2020} are sufficient for training state-of-the-art grasp detection networks such as \cite{Zhang.2021, M.Sundermeyer.2020,Mahler.2019}, they are not applicable to recent multi-modal sensor fusion approaches \cite{Gou.03.03.2021,Cao.2021, P.Hopfgarten.2020, Zhao.28.02.2020}. Besides this limitation, order picking systems usually depend on an additional upstream object detection. %based on color images. 
%In order to generate large-scale training data for these systems from simulation, high-quality 3D scans of objects are required. %
Existing datasets containing textured objects \cite{B.Calli.2015,Liu.23.04.2021,Fang.62020} are often limited to the household domain, available in small numbers and represent only a small subset of possible objects in warehouse or industry settings or do not contain real world scans \cite{shapenet2015}.

\medskip
%\subsection{Parallel-Jaw Grasp Labels and Datasets}
\noindent
\textbf{Parallel-Jaw Grasp Labels and Datasets:} Robotic grasping datasets can be categorized by the way its grasp labels are generated. %\cite{Kalashnikov.27.06.2018} and \cite{Levine.2018} performed extensive grasp attempts over several weeks with real hardware in order to train an agent predicting suitable grasp poses \cite{Kalashnikov.27.06.2018}, or even together with learning the robot’s kinematic \cite{Levine.2018}.
When only considering grasps with four degrees of freedom (DoF),
%where the approach direction of the gripper is aligned with the camera’s principal axis,
an antipodal grasp can be represented in image space by an oriented bounding box \cite{YunJiang.2011,H.Zhang.2019} (circle for a suction cup). By transferring the scene and grasp label generation into simulation, \cite{A.Depierre.2010} was able to increase the dataset size by a factor up to 50 with regard to \cite{YunJiang.2011}. With increasing numbers of objects in the scene and complex shape the advantage of 6 DoF grasps becomes more remarkable. Since generating such grasp labels in $SE(3)$ can become very tedious% when trying to cover a wide range of possible grasps in SE(3) for a given object
, recent 6 DoF datasets rely on automatic sampling schemes for grasp candidates. The generation of these labels is either based on analytical models such as antipodal-based samplers used in \cite{Fang.62020,Zhang.29.04.2021} or physics simulation which combines analytical sampling with physics simulators \cite{Mahler.2019,C.Eppner.2020}. For an in-depth overview % over current sampling methods
we refer to \cite{Eppner.11.12.2019}.

\medskip
%\subsection{Vacuum Grasp Labels and Datasets}
\noindent
\textbf{Vacuum Grasp Labels and Datasets:} Despite the wide deployment of vacuum-based robotic handling systems in automation the generation of reliable vacuum grasp labels based on the object's local shape remains an open field of robotic research, though offering a high potential for energy savings \cite{Gabriel.2020}. 
In \cite{Zeng.2019} appropriate vacuum contact points are sampled by hand. Due to the high-labeling costs their dataset is with 1837 annotated images small in scale.
%, but still shows promising results on their considered use-case evaluated against \cite{Mahler.20.09.2017}.%
Instead of labelling appropriate areas where vacuum seal can be applied from human experience, Dex-Net 3.0 \cite{Mahler.20.09.2017} models the suction cup as spring system and aims to find suitable suction points on 3D meshes’ surfaces in simulation. The spring model originates from \cite{Provot1995}, but instead of dynamically simulating the deformation of the suction cup over time as in \cite{Bernardin.2019}, they simplify the problem by only considering the quasi-static projection of the cup on to the object’s surface and evaluating the geometric deformation of each spring individually. 
SuctionNet1-Billion \cite{Cao.2021} simplifies \cite{Mahler.20.09.2017} cup model and replaces the original binary scoring by a continuous sealability score.
%SuctionNet1-Billion \cite{Cao.2021} builds upon \cite{Mahler.20.09.2017} spring model, but concludes that it is sufficient to only keep the perimeter springs of the original model. In addition, they replace the original binary vacuum seal scoring by a continuous scoring and apply their model to the two-step annotation process from \cite{Fang.62020}.
In contrast to \cite{Mahler.20.09.2017} their resulting dataset SuctionNet-1Billion has real RGBD  data and  multiple objects per scene. However, using the semi-automatic annotation process and data from \cite{Fang.62020} requires manually annoting the objects pose once per scene introducing label inaccuracies, and limiting the items' arrangement and the overall scene number.
%as well as leading to small annotation errors.
%introduces annotation inaccuracies.
%limits their scene number to 190 and leading to %since % and although presenting an innovative way to generate the labels, 
%the objects' poses have to labeled manually once per scene resulting in small 
%errors and limiting the arrangement of the items% in the scene.
In \cite{Zhang.2021} an adaption of \cite{Mahler.20.09.2017} model is proposed using a Grasp Wrench Space together with a weighting-scheme for the model's mass points. Despite beeing large in scale, their provided dataset is with colorless 11x11 pointclouds highly customized to their CNN-based network architecture for cloosed-loop grasping.

%\subsection{Object Detection and Scene Layout Datasets}
%There exist a lot of common object detection and segmentation datasets \cite{coco_dataset, cityscapes, imagenet} that provide precise object labels, but object detection in bin picking application sits on a completely different landscape, largely due to close-up views, diverse object poses, and occlusions. 

\medskip
\noindent
\textbf{Object Detection and Scene Layout Datasets:}  Existing object detection datasets in bin picking \cite{danielczuk_2019, Back.23.09.2021} emphasize scales and number of classes to boost model performance. These datasets do not focus on the unique challenges (described in \ref{sec:intro}) in object detection for bin picking.
%In particular, the challenges related to scene layouts are not addressed.
%Objects can be randomly positioned inside the bin. 
In highly cluttered scenes items can overlap or are wedged into each other. Simply inferring grasps 
%for the given sensor information
without considering the underlying scene layout might result in unsuccessful grasp attempts or even damaged objects. 
%\subsection{Manipulation Order}
Recent work attempts to tackle this problem by trying to learn the manipulation order for a picking system \cite{Kleeberger.12.01.2021,Zuo.2021}. However, there are currently only a few datasets available providing the required scene layout information \cite{H.Zhang.2019,Zhang.29.04.2021}. VMRD \cite{H.Zhang.2019} 
provides a dataset of over 5k scenes manually annotated. %with manually labelled top-down grasps. 
%The scenes are created by a human operator and contain beside the layout information image labels for 2D bounding box object detection and 4 DoF grasp detection.
REGRAD  \cite{Zhang.29.04.2021} uses simulation to increase the dataset size and provides 6 DoF parallel-jaw grasps. UOAIS \cite{Back.23.09.2021} provides amodal instance segmentation masks to reason about grasp scenes.
%However, it is located at household-domain and does not address our challenges related to object detection and ambidextrous bin picking . 
%In \cite{Back.23.09.2021} the occlusion of objects (amodal segmentation mask) is predicted to reason about the object layout and manipulation order. 
Despite the high relevance of these works, their datasets do not provide comprehensive modalities, grasping types, object types, as well as layout labels at the same time to address the complex bin picking problem. %is limited to the task of amodal instance segmentation and does not target other aspects of bin picking.

\medskip
%\subsection{Pose Estimation and Keypoint Detection}
\noindent
\textbf{Pose Estimation and Keypoint Detection:} Although grasping detection methods are model-free, fast and efficient,
%Earlier work in grasping detection often contains two parts: 3D model template matching and pose estimation. These methods are slowly replaced by  grasping detection methods. 
object pose estimation is still crucial for scene layout understanding or precise placement of picked items. In SynPick \cite{SynPick} a dataset for object pose tracking in dense bin clutter is proposed  simulating object-gripper interaction over time.
%and provide object poses and layout graphs for every time step.
Generating a 3D model for each real-world object is cost-inefficient, and unflexible. Object keypoints is an inexpensive way to describe object shapes and poses in grasping. In fact, many grasping detection and pose estimation works are based on object keypoints \cite{Li.2021, Z.Hu.2021}, but the dataset used in these works are limited in terms of scale and class diversity.

\section{Method}
     \vspace{-4mm}
%   \begin{figure}[ht]
%    \centering
%    \includegraphics[width=\columnwidth]{Figures/Grafik_Dataset_Generation_Pipeline.png}
%    \caption{Data generation pipeline for MetaGraspNet.}
%     \label{dataset_pipeline}
%   \end{figure}

   \begin{figure}[h]
    \centering
    \includegraphics[width=1.0\columnwidth]{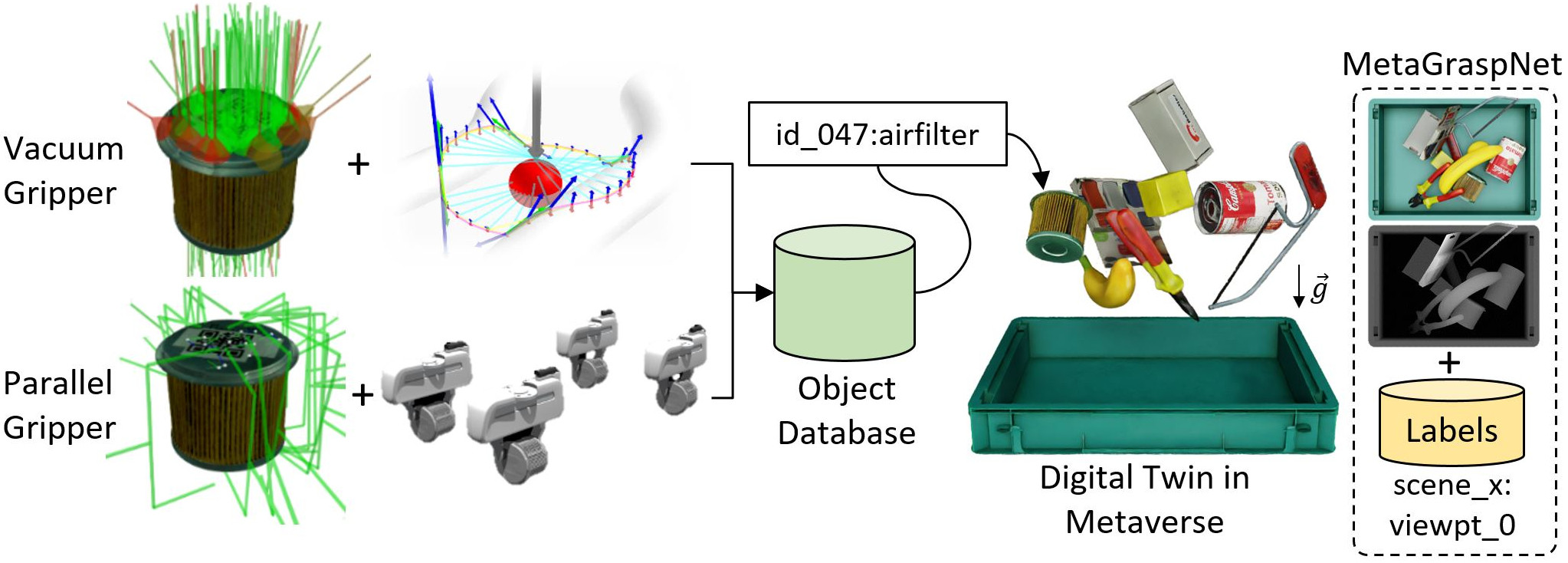}
        \vspace{-8mm}
    \caption{MetaGraspNet data generation pipeline.}
    \label{data_pipeline_overview}
    \vspace{-4mm}
   \end{figure}

The proposed method to generate MetaGraspNet can be divided into three steps: putting together a diverse item set, sampling ambidextrous grasp labels for each object individually, generating bin scenes together with rich annotations in the metaverse (see Fig. \ref{data_pipeline_overview}).

\subsection{Custom Object Dataset and Novel Object Testset}

%As pointed out in \ref{objects_set} only a small subset of existing object datasets contains high-quality textured meshes needed for generating photorealistic scenes in simulation.
In order to cover a broad area of possible use cases, we extended existing object sets with custom scans regarding the following criteria: parallel and vacuum grasp capability, transparency, reflectiveness, dimensions, industry/warehouse domain, deformability, texture, weight and fragility. 
%For an in-depth overview and rating of our proposed objects w.r.t. to the aforementioned criteria please refer to T in the appendix.%
%While scanning the objects for \cite{B.Calli.2015} required a lot of engineering \cite{Singh.52014}, 
The advancement of affordable  consumer-grade  and precise 3D scanner hardware (SHINING 3D EinScan-SP) allows to generate custom 3D models for individual use-cases. For our work we chose a subset of 33 high-quality meshes from \cite{Cao.2021} being part of YCB object set \cite{B.Calli.2015}, 4 from \cite{Liu.23.04.2021}, scanned 36 objects by ourselves and remodeled 9 in CAD software when scanning was not possible.

\begin{table}[htbp]
\tiny
\vspace{-3mm}
\caption{Novel Object List}
\vspace{-5mm}
\label{table:novel_object}
\begin{center}
\begin{tabular}{l cccc}
%\begin{tabular*}{\columnwidth}{@{\extracolsep{\fill}} l cccc}
        \toprule
        Class & Non-Convex & Black & Varying Shape & Transparent \\ 
        \midrule
         Pear       &               &  &  \checkmark & \\ 
         
         Mug        &               &  &   & \\

         Power drill      &   \checkmark  &  &  & \\
        
         Crayon box &               &  &  & \\
        
         Black clamp& \checkmark    & \checkmark & \checkmark & \\
        
         Black marker&              & \checkmark & \checkmark &\\
        
         Wire       & \checkmark    & & \checkmark & \\
        
         Wire in a bag &            & & \checkmark & \checkmark\\
        
         Wineglass  &  \checkmark   & & & \checkmark\\
        
         Eyeglasses &  \checkmark   & \checkmark & \checkmark & \checkmark\\
        \bottomrule
\end{tabular}
\end{center}
\vspace{-4mm}
\end{table}

Obtaining accurate 3D models for all objects is challenging and time consuming. Objects with existing 3D models can also be defective or deformed due to physical damage and alter from its rigid model. Therefore, it is crucial to evaluate object and grasping detection models with novel objects (objects that have never been seen before) to ensure the functionality of scene understanding and grasping beyond existing classes. We construct a novel object testset %for evaluating object and grasping detection methods 
on the following properties: convex/non-convex shape, transparency, varying shape, and black color. 
Non-convex shaped objects are harder to detect due to center of mass being outside of objects' body. Objects with varying shape can be challenging to detect in their entirety. Transparency and black color can make the value in depth and point cloud sensor incorrect or invalid. The list of novel objects and their properties are shown in Table \ref{table:novel_object}.

%\begin{table}[htbp]
%\tiny
%\caption{Table description for novel objects}
%\vspace{-4mm}
%\label{table:novel_object}
%\begin{center}
%\begin{tabular}{|c|cccc|}
%        \hline
%        Class & Non-convex & Black & Varying shape & Transparent \\ 
%        \hline
%         Pear       &               &  &  \checkmark & \\ 
%        \hline
%         Drill      &   \checkmark  &  &  & \\
%        \hline
%         Crayon Box &               &  &  & \\
%        \hline
%         Black clamp& \checkmark    & \checkmark & \checkmark & \\
%        \hline
%         Black Marker&              & \checkmark & \checkmark &\\
%        \hline
%         Wire       & \checkmark    & & \checkmark & \\
%        \hline
%         Wire in a bag &            & & \checkmark & \checkmark\\
%        \hline
%         Wineglass  &  \checkmark   & & & \checkmark\\
%        \hline
%         Eyeglasses &  \checkmark   & \checkmark & \checkmark & \checkmark\\
%        \hline
%\end{tabular}
%\end{center}
%\end{table}

\subsection{Parallel-Jaw Grasps Sampling Strategy} \label{pj_grasp_sampling}
In \cite{C.Eppner.2020}, the authors demonstrate the effectiveness of having a combination of antipodal sampling and physics simulation. Our proposed parallel-jaw grasps sampling method is inspired by their approach, however we expand it by a robust sampling strategy and an improved dynamic collision check.
For each object in our dataset we generate up to 5k antipodal grasps $G_j$ by sampling finger-object contact points $c_i$ evenly distributed over the mesh's surface. For each contact point $c_i$ we sample $k{=}1\ldots N,N{=}5$ antipodal \cite{Murray.op.1994} grasp attempts $c_{i,k}$ with random deviation in approach direction and translation. 
%A contact is considered successful if for both contact points the fingers' closing vector lies within a predefined friction cone and the distance between them is less than a predefined gripper opening width. 
The robust antipodal score  $s_{antip.,i}$ for a contact point $c_i$ is then defined as the number of successful samples divided by the number of total samples $N$. To obtain grasp poses in $SE(3)$ we sample for each successful contact point $s_{antip.,i} {>} 0$ up to $l{=}1\ldots L$ gripper poses by rotating it around the fingers’ closing direction. A grasp $  G_j{=}G_{i,k,l}$ is considered successful if the gripper does not collide with the object  and we assign it $s_{pj,anal.,j} {=} s_{antip.,i}$.
%For collision detection we modify the provided collision meshes from franka emika in such a way that the opening width can be adapted to the current grasp configuration resulting in more reachable grasps for objects with complex shape.
In the next step, each successful grasp $G_j$ is executed multiple times in a physics simulation in IsaacGym \cite{Makoviychuk.24.08.2021}. Again we extend the idea of robust sampling into simulation: Each grasp $G_j$ is simulated with different mass density factors and friction coefficients. Similar to \cite{C.Eppner.2020} we perform an upward and rotating gripper movement and assume a grasp is successful if the object is still in contact after execution. The robust simulation score $s_{pj,sim.,j}$ is then defined as the fraction of successful grasps divided by the total number of attempts.

%Sampling 25k grasps and then executing up to 20k grasps per object in simulation is computational expensive. Though being able to simulate over 900 grasps in parallel on a 2080 Ti GPU, generating all grasp labels for our custom dataset took over a week.

\subsection{Vacuum Seal Sampling Strategy} \label{vacuum_grasp_sampling}
\begin{figure}
    \centering
    \includegraphics[width=1.0\columnwidth]{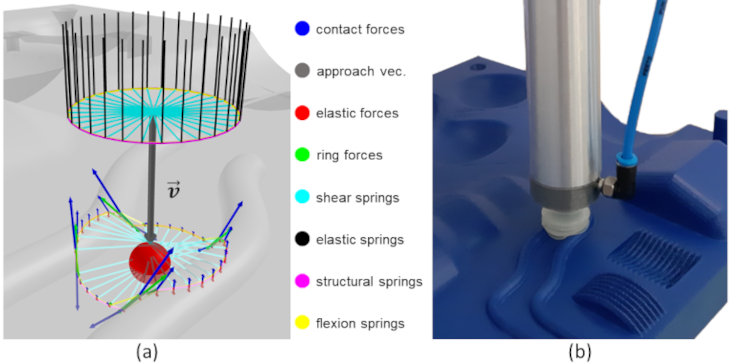}
    \caption{(a) Our proposed spring-mass system in simulation  before and after projection for a failed vacuum seal attempt. (b) We optimize for the resulting model parameters performing grasps on our calibration board.}
     \label{suction model}
     \vspace{-6mm}
   \end{figure}

In order to minimize the sim-to-real gap for vacuum sealability we propose a new physics-based vacuum suction cup model. Within the model, we adapt the projection idea of \cite{Mahler.20.09.2017} due to its universality and efficiency, but introduce a new spring-mass structure (see Fig. \ref{suction model}a). Moreover, in contrast to \cite{Cao.2021} our proposed model computes the actual forces within the spring-mass model and detects leakages between suction cup and object by analyzing the resulting force vector for each mass point locally.

For the projected spring-mass system, we assume mechanical equilibrium both over all mass points (globally) and at each individual mass point $m_i$ (locally). As forces we consider ring forces $\vec{f}_{r,i}$ obtained from the spring-mass structure, contact forces $\vec{f}_{p,i}$ due to the pressure difference  and elastic forces $\vec{f}_{e,i}$ resulting from the compression of the suction cup. While the ring forces $\vec{f}_{r,i}$ can be calculated directly via the deformation of the projected spring structure, we use the global equilibrium in Eq. (\ref{eq1}) to calculate the forces $\vec{f}_{e,i}$ introduced by the elastic springs:

%When projected onto the object's surface we assume a mechanical equilibrium summed up over all mass points $m_i$ between ring forces $\vec{f}_{r,i}$ in the suction cup, contact forces due to the pressure difference $\vec{f}_{p,i}$ and elastic forces $\vec{f}_{e,i}$ introduced by the elastic springs.

\begin{equation}
\sum_{i=0}^{n}\vec{f}_{r,i} + \sum_{i=0}^{n}\vec{f}_{p,i} + \sum_{i=0}^{n}\vec{f}_{e,i} = 0\label{eq1}
\end{equation}

With $\sum_{i=0}^{n}{\vec{f}_{r,i}} = 0$ and by assuming that $f_{e,i} = k_e \Delta l_i$ only act in approach direction $\vec{v}$, one can rewrite Eq. (\ref{eq1}) as a function of the relative elastic spring deformation $\Delta l_i$:
\begin{equation}
    F_p = \| \sum_{i=0}^{n}\vec{f}_{p,i} \|  = \| \sum_{i=0}^{n}  \vec{f}_{e,i} \| = k_e \sum_{i=0}^{n} \Delta l_i \label{eq2}
\end{equation}
Since in general $\Delta l_i$ cannot be directly examined from the projection of the spring-mass system, we express it as the difference between the maximum compression $\Delta l_{max}$ and the difference in length of each elastic spring $l_i$ with regard to the maximum compression: $\Delta l_i = \Delta l_{max} - l_i$.

Thus, Eq. (\ref{eq2}) can be solved for $\Delta l_{max}$ and we obtain the elastic force as a function of $l_i$ which can be directly inferred from the geometric projection of the mass points, whereby the vacuum force can be calculated using $F_p = \Delta p \pi r^2$:
\begin{equation}
    \vec{f}_{e,i} = f_{e,i} \cdot \vec{v}= k_e \cdot ( \frac{\frac{F_p}{k_e} + \sum l_i}{n} - l_i) \cdot \vec{v}
\end{equation}
Knowing the elastic forces $\vec{f}_{e,i}$ and the ring forces $\vec{f}_{r,i}$ for each mass point, we can compute the contact forces $\vec{f}_{p,i}$ using the local equilibrium. The vacuum seal is then checked by analyzing the resulting force direction for each mass point individually. We assume that the seal between cup and object breaks when the resulting force vector for each mass point becomes greater than zero in local normal direction, lifting the cup from the surface. 

By assuming that $\Delta l_{max}$ and the ratio between the calculated forces are independent of $n$ and all springs have same radius, we can reduce the number of parameters in our model from five to two. We perform 145 real world experiments in a robotic cell with a custom 3D printed vacuum seal board (see Fig. \ref{suction model}b) and use Bayesian method \cite{Mockus77} to optimize for the resulting two parameters. For every grasp attempt we record the seal by measuring the tear-off-force with the robotic arm Franka Emika Panda and the spring deformation for the same grasp configuration in simulation.

\subsection{Scene Generation and Object Labels}
Instead of generating and labelling scenes manually \cite{Zeng.2019, H.Zhang.2019} or semi-automatically \cite{Fang.62020, Cao.2021} we take inspiration from the recent rise of metaverse and create bin scenes completely in NVIDIA Isaac Sim \cite{NVIDIADeveloper.2019}. In a digital twin of a real-world bin picking scenario, we let objects drop randomly into the tote. The realistic physics-based interaction between the objects $k , k{=}1\ldots N$ and the bin assures that object layouts are realistic and physically accurate. Each scene is captured from 37 different camera viewpoints with alternating lightning conditions. Using path-tracing as rendering setting enables us to capture realistic light and shadow configurations as well as photo-realistic rendering of materials such as glass, plastic or metal. 
For each viewpoint all the individual objects’ parallel $G_{pj,k,j}$ \ref{pj_grasp_sampling} and vacuum suction grasps $G_{sc,k,j}$ \ref{vacuum_grasp_sampling} are checked for visibility and collision with other objects or the tote when approaching the scene and performing the grasp. A good grasp not only depends on the local object's surface (see \ref{pj_grasp_sampling} and \ref{vacuum_grasp_sampling}), but is also highly affected by the wrenches applied to the gripper contact. %For the vacuum grasp only the seal was considered, therefore 
For each vacuum grasp the wrench is computed around all three contact axes and scored similar to \cite{Cao.2021} $s_{sc,sim,j}\in[0,1]$ while taking into account the object's orientation and its center of mass. Though being implicitly considered in the $s_{pj,sim}$ (see \ref{pj_grasp_sampling}) we also specify an explicit wrench score $s_{pj,soft,j} \in [0,1]$ around the finger's closing direction for each parallel jaw grasp $j$ (soft-finger contact \cite{Murray.op.1994}). 

\begin{figure}
\centering
\includegraphics[width=\columnwidth]{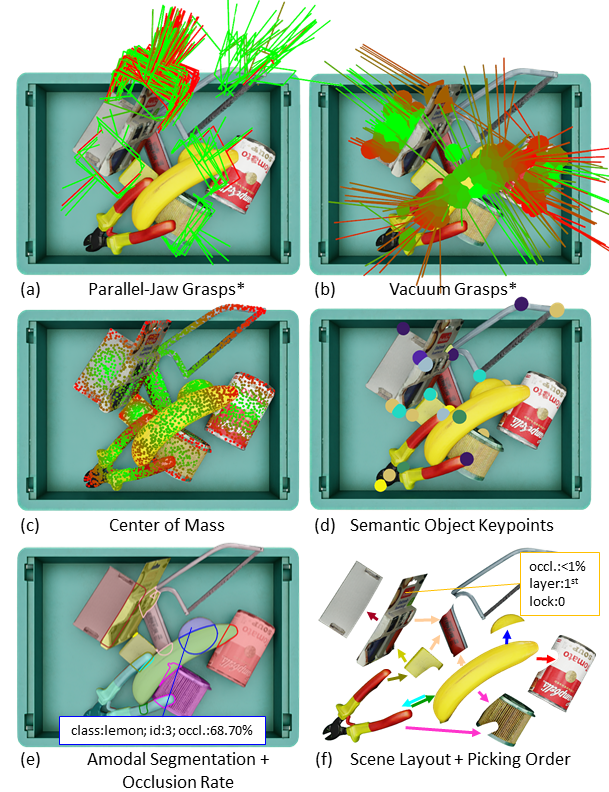}
%\vspace{-4mm}
\caption{Amdidextrous grasp and scene labels are provided for each viewpoint ($^{*}$only subset visualized).}
 \label{dataset_labels}
 \vspace{-6mm}
\end{figure}

Besides grasp labels, we provide extensive object labels for each viewpoint including amodal segmentation masks and occlusion rate, semantic keypoints and center of mass distribution heat maps  (see Fig. \ref{dataset_labels} (a-e)). 
We define the amodal segmentation mask as a tuple of pixel-wise occlusion masks $M_{occl., k}$ for each object instance $k$ in the scene. The occlusion score $s_{occl.,k} \in [0,1)$ is then defined as the quotient of occluded $M_{occl., k}$ and total object surface area $M_{total, k} = M_{occl.,k} \cup M_{vis., k}$. 
%Its computation is based on visible objects segmentation masks by hiding and showing objects in the scene.
Object keypoints are manually labeled on 3D object models and represent joints or surface centers. We transform them into the scene and perform ray-tracing to check for visibility.
%with the intrinsic camera parameters and check for collision. 
A keypoint $x_{key,k}~=~[id_{sem}, (x,y), id_{class}, id_{instance}]$ is defined as a tuple of image coordinates $(x,y)$, its unique semantic $id_{sem}$, class category $id_{class}$, and instance $id_{instance}$ id.

%The center of mass heatmap is done via keypoint detection as well. In a pre-processing step we sampled for each object $L=1000$ surface points and scored them with values $id_{sem,k,l} = s_{com,l} \in [0,1]$ based on the distance to the center of mass. (TODO add amodal instance segmentation masks)

\subsection{Scene Layout Label and Scene Difficulties}
\label{sec:layout-label}

\begin{table}[htbp]
\tiny
\vspace{-2mm}
\caption{5 Difficulty Levels}
\label{table:level_description}
\begin{center}
\vspace{-4mm}
\begin{tabular}{lcccc}
        %\hline
        \toprule
        Level & Layer Limit & Occlusion Limit (\%) & Complete Object & Unique Class\\ 
        %\hline
        \midrule
         1 & 2 & 5 & \checkmark & \checkmark\\ 
        %\hline 
         2 & N/A & N/A & \checkmark & \checkmark\\
        %\hline
         3 & N/A & N/A &  & \checkmark\\
        %\hline
         4 & N/A & N/A & \checkmark & \\
        %\hline
         5 & N/A & N/A &&\\
        \bottomrule
        %\hline
\end{tabular}
\end{center}
\vspace{-6mm}
\end{table}

We propose two additional labels to characterize the scene layouts.
The first label is a matrix storing the relation between each pair of objects, providing a comprehensive layout representation.
To construct the relation matrix, we define three types of relationship for a pair of object $A$ and $B$. 
If $A$ is occluding $B$, we define the relationship between $(A, B)$ as positive, with a numerical value of 1.
If A is occluded by B, we define the relationship between $(A, B)$ as negative, with a numerical value of -1.
If $A$ and $B$ have no direct relationship or $A=B$, we define the relationship between $(A, B)$ as neutral, with a numerical value of 0.
Based on these definitions, for a layout with $N$ objects, we create a relation matrix with $N{\times} N$ elements, where element $(i, j)$ in the matrix is the relationship between object $i$ and object $j$.

The second label provides a simpler layout description in line with the bin picking task.
%For each object in the environment, we want to answer to the following question.
%How many other objects are on top of the current object that need to be moved away before picking?
To better understand the order in which objects must be grasped, we create a directed graph to represent each layout.
%Each node represents an object in the layout and each edge represents an obstruction relationship where the parent object is covering the child.
%From this representation, we can see what objects and how many objects are occluding the same object.
As robots pick objects sequentially, occluded objects will be revealed entirely once the objects on top of them are picked.
%Therefore, it is not necessary to evaluate occluded objects that are at the bottom of the scene.
%Sometimes, the occlusion between objects is small enough that can be ignored, so objects occluded by only a single other objects are also important to be evaluated. 
Given this, we categorize each object in a layout into 3 different layers.
Top layer contains objects that are clear of any obstructions.
Secondary layer includes objects that are covered by only a single other object.
Others layer includes the rest of the objects.
In some cases, there could be groups of interlocked objects.
Interlocked objects that are being directly covered by only one object would be considered to be within the secondary layer.
An example of a environment of objects from the top down view and the resulting graph can be seen in Fig. \ref{dataset_labels} (f).

%\subsection{Layout-based Difficulty Levels}
%\label{difficulty_levels}
%While per category object detection metrics can measure performance on specific categories of objects, 
A difficulty rating for each scene would allow us to better understand how the model would perform under different environment conditions.
We label images according to 5 different levels of difficulty.
Those levels are defined by 4 different characteristics: Number of layers, occlusion percentage, instance completeness, and class uniqueness.
Instance completeness refers to if a single object instance is visually crosscut into multiple segments due to occlusion.
%In such case, we refer to this kind of objects as incomplete objects, and refer to objects without visual crosscuts as complete objects.
%Incomplete objects often 
This often causes object over-detection or over-segmentation in objection detection and segmentation methods. %, and thus it is a good characteristic to test object detection and segmentation models.
Class uniqueness is if all objects in an image belong to different categories, or are visually distinct from each other.
This characteristic evaluates models on distinguishing objects with similar visual features while clustered.
The first two difficulty levels will be primarily concerned with understanding how a model deals with different levels of occlusion and layers.
%The layer limit for level 1 difficulty is set to 1, and the occlusion limit is set to 5\% empirically.
%A scene of difficulty level 1 should include no more than 2 layers (top, secondary), and contain a maximum occlusion score of ****something****
The next three levels measures the model's ability to correctly label object instances. Level 3 includes incomplete objects in an image, and level 4 includes non-unique objects. Level 5 includes both incomplete as well as non-unique objects.
The properties of all difficulties levels are shown in Table \ref{table:level_description}.

\section{Dataset Details}
The proposed MetaGraspNet benchmark dataset contains 217k RGBD images % respectively colored point clouds, 
with 5884 different
scenes and 82 different objects from household and logistics domain. 
Along with the RGBD images, camera parameters are provided for generating point clouds.
%Techniques such as domain randomization has shown to be effective for minimizing the gap between simulation and real world. 
%The objects are dropped randomly into the bin, representing a universal bin picking use case in intralogistics. 
%A scene is a single arrangement of objects in the bin captured from 37 camera viewpoints. 
All labels are provided in the respective camera coordinate system for each viewpoint, arranged in a hemisphere around the bin.
%Although we do not consider techniques such as domain randomization, we provide the option to generate custom scenes with heavy randomization of camera viewpoint, lighting, materials and textures.

Besides a large-scale synthetic dataset, a smaller real-world evaluation and novel object test set is provided. It contains out of over 580 bin scenes equally distributed over the proposed five layout difficulty levels. Each scene is captured with a high performance 3D vision system based on time-coded structured light (Zivid Two) mounted at the robot's endeffector from top view and three randomly sampled poses out of the bin hemisphere. Annotations are pixel-wise with semantic and instance segmentation masks, object layout as well as vacuum and parallel-jaw grasp labels. In total, 2.3k RGBD images of real world bin scenes containing over 9.7k items out of 76 classes are provided.
\section{Experiments}
%All grasp experiments are conducted on a Franka Emika Panda robot equipped with a Zivid Two vision system and a custom 3D printed endeffector.
\begin{figure}
\centering
\includegraphics[width=0.9\columnwidth]{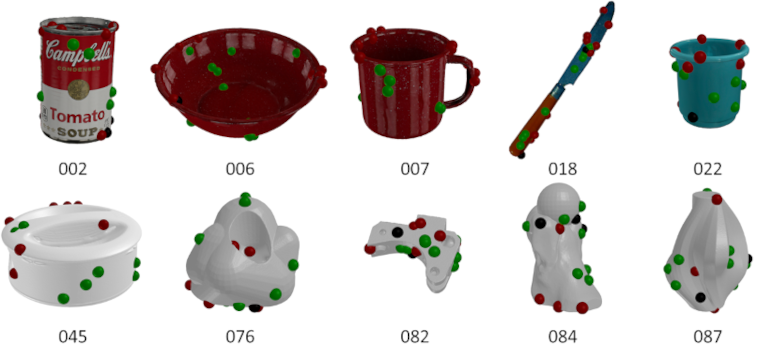}
\caption{Overview over items used for evaluating our proposed vacuum seal model. (top row): household objects \cite{B.Calli.2015}, (bottom row): complex adversarial objects \cite{Mahler.20.09.2017}. Correct predictions are visualized with green (true positive) and red (true negative) spheres. Black spheres mark false predictions (false positive and negative).}
 \label{item_seal_evaluation}
\vspace{-3mm}
\end{figure} 
\begin{figure}
\centering
\includegraphics[width=\columnwidth]{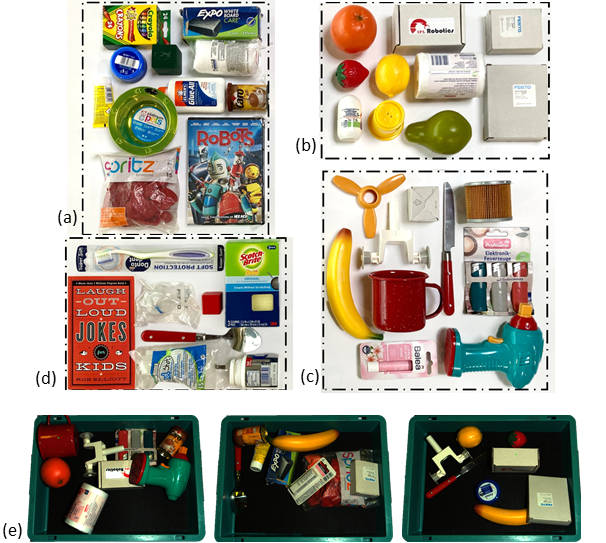}
\caption{Items used for vacuum bin picking experiments separated in four subsets: (a) unseen easy, (b) seen easy, (c) seen hard, (d) unseen hard. \\3 out of 40 test scenarios are shown in (e).}
 \label{item_sets}
 \vspace{-6mm}
\end{figure} 
\subsection{Vacuum Grasp Labels}\label{exp_grasp_labels}

The proposed physics based sealability model is evaluated with real world grasp experiments. In detail, answering the following questions are of interest: 1) Does the model generalize to different cup materials and dimensions? 2) How accurate does the model perform on real world objects and compared to current state-of-the-art methods?

By performing four times 60 grasps on our custom board on a separate test split with different suction cups, it can be shown that the proposed method generalizes well to common cup materials, sizes and shapes (see Table \ref{evaluation_vacuum_seal_labels} \textit{board} experiments).
To evaluate the performance on real-world objects, experiments on household \cite{B.Calli.2015} and 3D printed adversarial objects \cite{Mahler.20.09.2017} both used in \cite{Cao.2021} (see Fig. \ref{item_seal_evaluation}) are performed. For each object its 6D pose in the robotic cell is registered by choosing corresponding keypoints between mesh and sensor pointcloud and refining the first estimate with an ICP registration pipeline provided by open3d \cite{Zhou2018}. In total, by performing 200 grasps  (10 positive and 10 negative predicted seal per object) for each cup model it can be shown that the proposed  model achieves a very high performance on real-world objects and is able to generalize to different cup sizes as well.
(see Table \ref{evaluation_vacuum_seal_labels} \textit{real} experiments). 

In order to benchmark our model against related work in the field, the provided sealability score of SuctionNet-1Billion \cite{Cao.2021} is compared with our model's prediction and a reimplementation of Dex-Net 3.0 \cite{Mahler.20.09.2017}. For this experiment only controversial contact points are considered in order to emphasize the difference between these methods (11.0\% for \cite{Mahler.20.09.2017} and 22.4\% for \cite{Cao.2021} of total amount). A point is considered controversial, if its given seal score is below 0.2 \cite{Cao.2021} while our method predicts a successful vacuum seal, or respectively if the score is above 0.8 and our methods predicts a failed seal. As shown in Table \ref{evaluation_vacuum_seal_labels_suc_dex} our proposed physics based suction cup models outperforms both methods by a large margin. Looking at the results for the experiments with the 30mm diameter suction cup, once again the robustness of our model with regard to cup dimension changes can be confirmed.

\begin{table}[htbp]
\begin{threeparttable}
\tiny
\caption{Performance of our proposed vacuum sealability model evaluated on test board and real world objects}
\begin{center}
\begin{tabular*}{\columnwidth}{@{\extracolsep{\fill}}lcccccc}
\toprule
Experiment & Material, \diameter, Conv.\tnote{a} & Prec. (PPV)  & NPV  & Sens. & Spec. & Acc.\\
\midrule
board & Silicon, 20mm, 3.5\tnote{1} &0.98 & 0.87 & 0.96 & 0.93 & 0.95\\
board & NBR, 20mm, 3.5\tnote{2} &0.95 & 0.75 & 0.91 & 0.86 & 0.90\\
board& Silicon, 20mm, 1.5\tnote{3} &0.90 & 0.92 & 0.98 & 0.69 & 0.90\\
board& Silicon, 30mm, 3.5\tnote{4} & 0.97 & 0.86 & 0.92 & 0.95 & 0.93\\
\midrule
real & Silicon, 20mm, 3.5\tnote{1} &0.92 & 0.82 & 0.84 & 0.91 & 0.87\\
real & Silicon, 30mm, 3.5\tnote{4} &0.87 & 0.92 & 0.91 & 0.88 & 0.89\\
\bottomrule
\end{tabular*}

%\smallskip
\scriptsize

%\multicolumn{7}{l}{\textit{PPV}: positive predictive value, \textit{NPV}: negative predictive value, \textit{Conv.}: bellows convolutions, \textit{Prec.}: precision}\\
%\multicolumn{7}{l}{ \textit{Sens.}: sensitivity, \textit{Spec.}: specifity, \textit{Acc.}: accuracy} \\
\begin{tablenotes}
%\textit{PPV}: positive predictive value, \textit{NPV}: negative predictive value, \textit{Conv.}: bellows convolutions, \textit{Prec.}: precision\\
%\RaggedRight
%\multicolumn{7}{l}{\textit{PPV}: positive predictive value, \textit{NPV}: negative predictive value, \textit{Conv.}: bellows convolutions, \textit{Prec.}: precision}\\
\item PPV: Positive Predictive Value; NPV: Negative Predictive Value; Conv.: Bellows Convolutions; Sens.: Sensitivity; Prec.: Precision; Acc.: Accuracy
\item[a] 1: ESS-20-CS (used for training);
         2: ESS-20-CN;
         3: ESS-20-BS;
         4: ESS-30-CS
%\item[b] \dots
%\item[c] \dots
\end{tablenotes}
%\multicolumn{5}{l}{$^{\mathrm{1}}$: ESS-20-CS (used for training), $^{\mathrm{2}}$: ESS-20-BS, $^{\mathrm{3}}$: ESS-20-CN, $^{\mathrm{4}}$: ESS-30-CS}\\
\label{evaluation_vacuum_seal_labels}
\end{center}
\end{threeparttable}
\vspace{-2mm}
\end{table}

%\begin{table}[htbp]
%\tiny
%\caption{Performance comparison of our proposed vacuum seal model against \cite{Mahler.20.09.2017} and \cite{Cao.2021}}
%\label{evaluation_vacuum_seal_labels_suc_dex}
%\begin{center}
%\begin{tabular}{|c|c|c|c|c|c|}
%\hline
%\textit{method} & \textit{Precision (PPV)}  & \textit{NPV} & \textit{Sensitivity} & \textit{Specificity} & \textit{Accuracy}\\
%\hline
%\cite{Mahler.20.09.2017}$^{\mathrm{1}}$ & 12/20 = 0.60  & 11/60 = 0.18 & 12/61 = 0.20 & \textbf{11/19 = 0.58} & 23/80 = 0.29\\
%\textbf{ours}$^{\mathrm{1}}$ &\textbf{49/60 = 0.82} & \textbf{8/20 = 0.4} &	\textbf{49/61 = 0.80} & 8/19 = 0.42 & \textbf{57/80 = 0.71}\\
%\hline
%\cite{Cao.2021}$^{\mathrm{2}}$ & 13/44 = 0.30  & 6/24 = 0.25 & 13/31 = 0.42  & 6/37 = 0.16 & 19/68 = 0.28 \\
%\textbf{ours}$^{\mathrm{2}}$ &	\textbf{18/24 = 0.75} & \textbf{31/44 = 0.70 } & \textbf{18/31 = 0.58} & \textbf{31/37 = 0.84} & \textbf{49/68 = 0.72}\\
%\hline
%multicolumn{6}{l}{\textit{PPV}: positive predictive value, \textit{NPV}: negative predictive value}\\
%\multicolumn{6}{l}{$^{\mathrm{1}}$: outer-\diameter{20mm} suction cup (Festo ESS-20-CS), $^{\mathrm{2}}$: outer-\diameter{30mm} suction cup (Festo ESS-30-CS)}\\
%\end{tabular}
%\label{suctionnet_evaluation}
%\end{center}
%\end{table}

\begin{table}[htbp]
\begin{threeparttable}
\tiny
\caption{Performance comparison of our proposed vacuum seal model against Dex-Net 3.0 \cite{Mahler.20.09.2017} and SuctionNet-1Billion\cite{Cao.2021}}
\label{evaluation_vacuum_seal_labels_suc_dex}
\begin{center}
\begin{tabular*}{\columnwidth}{@{\extracolsep{\fill}}lccccc}

\toprule
Method\tnote{a} & Precision (PPV)  & NPV & Sensitivity & Specificity & Accuracy\\
\midrule
\cite{Mahler.20.09.2017}\tnote{1} & 12/20 = 0.60  & 11/60 = 0.18 & 12/61 = 0.20 & \textbf{11/19 = 0.58} & 23/80 = 0.29\\
ours\tnote{1} &\textbf{49/60 = 0.82} & \textbf{8/20 = 0.4} &	\textbf{49/61 = 0.80} & 8/19 = 0.42 & \textbf{57/80 = 0.71}\\
\midrule
\cite{Cao.2021}\tnote{4} & 13/44 = 0.30  & 6/24 = 0.25 & 13/31 = 0.42  & 6/37 = 0.16 & 19/68 = 0.28 \\
ours\tnote{4} &	\textbf{18/24 = 0.75} & \textbf{31/44 = 0.70 } & \textbf{18/31 = 0.58} & \textbf{31/37 = 0.84} & \textbf{49/68 = 0.72}\\
\bottomrule
\end{tabular*}

%\smallskip
\scriptsize
\begin{tablenotes}
%\RaggedRight
\item PPV: Positive Predictive Value;
         NPV: Negative Predictive Value 
\item[a] 1: ESS-20-CS (\diameter 20mm); 4: ESS-30-CS (\diameter 30mm)
\end{tablenotes}

%\multicolumn{6}{l}{\textit{PPV}: positive predictive value, \textit{NPV}: negative predictive value}\\
%\multicolumn{6}{l}{$^{\mathrm{1}}$: outer-\diameter{20mm} suction cup (Festo ESS-20-CS), $^{\mathrm{2}}$: outer-\diameter{30mm} suction cup (Festo ESS-30-CS)}\\

\end{center}
\end{threeparttable}
\vspace{-4mm}
\end{table}

\subsection{Grasp Planning}

While previous work such as \cite{Mahler.2019} have shown the potential of synthetic depth data for training picking systems, the performance drop for RGBD based methods \cite{Cao.2021} from simulation to real world is still significant. Extensive experiments on a physical picking cell equipped with a Franka Emika robot arm and Zivid Two RGBD camera system can demonstrate that the proposed MetaGraspNet dataset is able to close the gap from simulation to real world for cluttered bin  scenes. 
%In detail, the provided on real RGBD images trained version of SuctionNet1-Billion\cite{Cao.2021} is evaluated against a version of their network optimized on our synthetic data. 
In detail, SuctionNet-1Billion \cite{Cao.2021} trained on its large-scale real-world dataset is evaluated against a version of \cite{Cao.2021} trained on the proposed synthetic MetaGraspNet database.
In total, 813 grasp attempts distributed over 40 cluttered bin layouts are analyzed. Each scene contains eight randomly sampled items arranged in random poses and partly stacked upon each other (see Fig. \ref{item_sets}). In order to avoid human bias, the manual scene creation and recreation for both networks is alternated. For the experiments, background was filtered out and a grasp was considered successful if the object was picked up and moved into another bin. After two failed grasps attempts per object and scene, a human supervisor removed the object. For evaluation the proposed metrics $R_{grasp}$, $R_{object}$ and $R_{mixture}$ from \cite{Cao.2021} are adapted. As shown in Table \ref{suctionnet_evaluation_detailed} \textit{all}, MetaGraspNet outperforms real-world training data in terms of total number of successful grasps $R_{grasp}$ and total number of autonomously cleared objects $R_{object}$. Looking at Table \ref{suctionnet_evaluation_detailed}, this observation is valid for known as well as unknown objects (see Fig. \ref{item_sets}). Only when it comes to successful first grasp attempts on cleared objects $R_{mixture}$, \cite{Cao.2021} trained on real data outperforms our method on seen objects.

\begin{table}[htbp]
\caption{Vacuum bin picking experiments with SuctionNet-1Billion trained on real data \cite{Cao.2021} and our synthetic data}
\vspace{-6mm}
%\begin{center}
\begin{threeparttable}
\label{suctionnet_evaluation_detailed}
\begin{tabular*}{\columnwidth}{@{\extracolsep{\fill}}l cccccc}
\toprule
Test Set & \multicolumn{2}{c}{$R_{grasp}$ (\%)} & \multicolumn{2}{c}{$R_{object}$ (\%)} & \multicolumn{2}{c}{$R_{mixture}$ (\%)}\\
\cmidrule{2-3}
\cmidrule{4-5}
\cmidrule{6-7}
\textbf{\textit{}} & \cite{Cao.2021}  & ours & \cite{Cao.2021}
 & ours & \cite{Cao.2021} & ours \\
\midrule
complete & 60.7 & \textbf{64.5} & 77.6 & \textbf{81.0} & \textbf{93.6} & 92.5\\
\addlinespace
seen & 69.5 & \textbf{71.2} & 82.8 & \textbf{87.8} & \textbf{97.6} & 88.8\\
seen\tnote{a} & 69.5 & \textbf{73.7} & 82.8 & \textbf{87.9} & \textbf{97.6} & 92.6\\
unseen & 56.8 & \textbf{57.1} & \textbf{75.1} & 73.6  & 91.2 & \textbf{96.5}\\
unseen\tnote{a} & 52.8 & \textbf{57.1} & 75.0 & \textbf{77.9}  & 81.4 & \textbf{84.5} \\
\bottomrule
\end{tabular*}

%\smallskip
\scriptsize
\begin{tablenotes}
%\RaggedRight
\item[a] intersecting set of \cite{Cao.2021} and our object set 
\end{tablenotes}
%\multicolumn{7}{l}{$^{\mathrm{*}}$: intersection for \cite{Cao.2021} and our object set} \\
%\end{center}
\vspace{-4mm}
\end{threeparttable}
\end{table}

\subsection{Object Detection}
Class-agnostic object detection and segmentation is one of the most important yet challenging tasks leading towards robust and reliable understanding of bin scenes. This task ensures the consistent model performance even if there are defective, or unseen objects in the bin. 
We use classic object detection and segmentation network Mask R-CNN \cite{maskrcnn} to evaluate the performance gap between our synthetic, real, and unseen datasets on RGB images for this task. % for class-agnostic item localisation inside the bin. 
%We treat all objects as 1 class to evaluate on unseen objects. The unseen objects defined in \ref{table:novel_object} are excluded from the synthetic and real dataset. 
We treat all objects as 1 class and exclude the unseen test objects (defined in \ref{table:novel_object}) from training.
%For the synthetic dataset, we used 80\% for training and 20\% for testing. 
The real dataset is used for either training or testing. The unseen dataset is used for testing only. We train our models on synthetic (Syn) or real (Real) dataset  for 37 epochs. We also evaluate a model trained on Syn and fine-tuned on Real (Syn+Real) for 7 epochs. We report our results in Bounding Box Average Precision (Box AP) and Segmentation Average Precision (Seg AP), as shown in \ref{table:object detection results}. From the results, we can see that models trained on synthetic and real dataset have very small performance gap on the unseen object test sets.  

\begin{table}[htbp]
\begin{center}
\begin{threeparttable}
\caption{Object detection performance gap between synthetic, real, and unseen data. }
\label{table:object detection results}

%\begin{tabular}{l c c c}
\begin{tabular*}{0.8\columnwidth}{@{\extracolsep{\fill}}l lcc}

%\hline
\toprule
        Train set & Test set\tnote{a} & Box AP & Seg AP\\ 
        %\hline
\midrule        
        %Syn & Syn & 62.6 & 54.5\\
        %\hline
        %Syn & Real & 45.5 & 40.0 \\
        %\hline
%\addlinespace        
        Syn & Unseen mix & 34.0 & 28.7 \\ 
        %\hline
        Real & Unseen mix & 34.0 & 32.1 \\ 
        %\hline
        Syn+Real & Unseen mix & 42.2 & 37.3 \\ 
        %\hline
\addlinespace        
        Syn & Unseen only & 12.2 & 9.5 \\ 
        %\hline
        Real & Unseen only & 9.1 & 10.7 \\ 
        %\hline
        Syn+Real & Unseen only & 13.2 & 12.4 \\ 
\bottomrule        
        %\hline 
\end{tabular*}
\scriptsize
\begin{tablenotes}
\item[a] Unseen mix: contains all the scenes with unseen objects

         Unseen only: contains only unseen objects.
\end{tablenotes}

\end{threeparttable}
\vspace{-6mm}
\end{center}
\end{table}

\section{CONCLUSIONS}
In this work, we introduced a large-scale comprehensive photo-realistic synthetic train and an extensive real-world evaluation and test dataset for robotic bin picking. Extensive robot experiments could show that the proposed vacuum grasp label generation method generalizes to different cup models and together with the proposed synthetic data generation pipeline outperforms real-world data for vacuum based bin picking in clutter. 
%Bin picking is highly complex and usually requires multi-stage approaches to solve it, 
With MetaGraspNet a data generation pipeline is introduced which addresses all vision-related aspects of bin picking and challenges future systems to take the next step towards scene understanding and ambidextrous manipulation.
%Looking at the potential of metaverses for customizing training data and the expected rapid progress, we designed the software modular, allowing to adapt for user-specific application areas, items or grippers. 
%With high interest we are following the recent advance in the field, and see great potential for future work by expanding our pipeline with highly-flexible items (e.g. textiles or plastic bags).

%or manipulation primitives (e.g. object pushing or toppling) which allow the robot to extend its interaction with the scene in order to find better grasps.

%%%%%%%%%%%%%%%%%%%%%%%%%%%%%%%%%%%%%%%%%%%%%%%%%%%%%%%%%%%%%%%%%%%%%%%%%%%%%%%%

%%%%%%%%%%%%%%%%%%%%%%%%%%%%%%%%%%%%%%%%%%%%%%%%%%%%%%%%%%%%%%%%%%%%%%%%%%%%%%%%

%%%%%%%%%%%%%%%%%%%%%%%%%%%%%%%%%%%%%%%%%%%%%%%%%%%%%%%%%%%%%%%%%%%%%%%%%%%%%%%%
%\section*{APPENDIX}

%Appendixes should appear before the acknowledgment.
\addtolength{\textheight}{-12cm}   % This command serves to balance the column lengths
                                  % on the last page of the document manually. It shortens
                                  % the textheight of the last page by a suitable amount.
                                  % This command does not take effect until the next page
                                  % so it should come on the page before the last. Make
                                  % sure that you do not shorten the textheight too much.

\section*{ACKNOWLEDGMENT}
We would like to thank NVIDIA for hardware support.

\bibliographystyle{IEEEtran}
\bibliography{refs}

\begin{thebibliography}{10}
\providecommand{\url}[1]{#1}
\csname url@rmstyle\endcsname
\providecommand{\newblock}{\relax}
\providecommand{\bibinfo}[2]{#2}
\providecommand\BIBentrySTDinterwordspacing{\spaceskip=0pt\relax}
\providecommand\BIBentryALTinterwordstretchfactor{4}
\providecommand\BIBentryALTinterwordspacing{\spaceskip=\fontdimen2\font plus
\BIBentryALTinterwordstretchfactor\fontdimen3\font minus
  \fontdimen4\font\relax}
\providecommand\BIBforeignlanguage[2]{{%
\expandafter\ifx\csname l@#1\endcsname\relax
\typeout{** WARNING: IEEEtran.bst: No hyphenation pattern has been}%
\typeout{** loaded for the language `#1'. Using the pattern for}%
\typeout{** the default language instead.}%
\else
\language=\csname l@#1\endcsname
\fi
#2}}

\bibitem{SynPick}
A.~Periyasamy, M.~Schwarz, and S.~Behnke, ``Synpick: A dataset for dynamic bin
  picking scene understanding,'' in \emph{Proc. IEEE Int. Conf. Automat. Sci.
  and Eng.}, 2021, pp. 488--493.

\bibitem{Zhang.29.04.2021}
H.~Zhang \emph{et~al.}, ``Regrad: A large-scale relational grasp dataset for
  safe and object-specific robotic grasping in clutter,'' \emph{IEEE Robot. and
  Automat. Lett.}, vol.~7, no.~2, pp. 2929--2936, 2022.

\bibitem{Cao.2021}
H.~Cao, H.~Fang, W.~Liu, and C.~Lu, ``Suctionnet-1billion: A large-scale
  benchmark for suction grasping,'' \emph{IEEE Robot. and Automat. Lett.},
  vol.~6, no.~4, pp. 8718--8725, 2021.

\bibitem{Mahler.2019}
J.~Mahler \emph{et~al.}, ``Learning ambidextrous robot grasping policies,''
  \emph{Science Robotics}, vol.~4, no.~26, 2019.

\bibitem{Kalashnikov.27.06.2018}
D.~Kalashnikov \emph{et~al.}, ``Scalable deep reinforcement learning for
  vision-based robotic manipulation,'' in \emph{Proc. Conf. on Robot Learn.},
  vol.~87, 2018, pp. 651--673.

\bibitem{Levine.2018}
S.~Levine, P.~Pastor, A.~Krizhevsky, J.~Ibarz, and D.~Quillen, ``Learning
  hand-eye coordination for robotic grasping with deep learning and large-scale
  data collection,'' \emph{Int. J. Robot. Res.}, vol.~37, pp. 421--436, 2018.

\bibitem{Zeng.2019}
A.~Zeng \emph{et~al.}, ``Robotic pick-and-place of novel objects in clutter
  with multi-affordance grasping and cross-domain image matching,'' \emph{Int.
  J. Robot. Res.}, Aug. 2019.

\bibitem{YunJiang.2011}
{Y. Jiang}, {S. Moseson}, and {A. Saxena}, ``Efficient grasping from rgbd
  images: Learning using a new rectangle representation,'' in \emph{Proc. IEEE
  Int. Conf. Robot. and Automat.}, 2011, pp. 3304--3311.

\bibitem{A.Depierre.2010}
A.~Depierre, E.~Dellandr{é}a, and L.~Chen, ``Jacquard: A large scale dataset
  for robotic grasp detection,'' in \emph{Proc. IEEE/RSJ Int. Conf. Intell.
  Robots and Syst.}, 2018, pp. 3511--3516.

\bibitem{shapenet2015}
A.~Chang \emph{et~al.}, ``{ShapeNet: An Information-Rich 3D Model
  Repository},'' \emph{arXiv:1512.03012}, 2015.

\bibitem{H.Zhang.2019}
H.~Zhang, X.~Lan, S.~Bai, X.~Zhou, Z.~Tian, and N.~Zheng, ``Roi-based robotic
  grasp detection for object overlapping scenes,'' in \emph{Proc. IEEE/RSJ Int.
  Conf. Intell. Robots and Syst.}, 2019, pp. 4768--4775.

\bibitem{Fang.62020}
H.~Fang, C.~Wang, M.~Gou, and C.~Lu, ``Graspnet-1billion: A large-scale
  benchmark for general object grasping,'' in \emph{Proc. IEEE/CVF Conf.
  Comput. Vision and Pattern Recognit.}, 2020, pp. 11\,444--11\,453.

\bibitem{B.Calli.2015}
{B. Calli}, {A. Singh}, {A. Walsman}, {S. Srinivasa}, {P. Abbeel}, and {A. M.
  Dollar}, ``The ycb object and model set: Towards common benchmarks for
  manipulation research,'' in \emph{Proc. Int. Conf. on Adv. Robot.}, 2015, pp.
  510--517.

\bibitem{xiang2018posecnn}
Y.~Xiang, T.~Schmidt, V.~Narayanan, and D.~Fox, ``Posecnn: A convolutional
  neural network for 6d object pose estimation in cluttered scenes,''
  \emph{Robot.: Sci. and Syst.}, 2018.

\bibitem{Liu.23.04.2021}
Z.~Liu \emph{et~al.}, ``Ocrtoc: A cloud-based competition and benchmark for
  robotic grasping and manipulation,'' \emph{IEEE Robot. and Automat. Lett.},
  vol.~7, no.~1, pp. 486--493, 2022.

\bibitem{Gou.03.03.2021}
M.~Gou, H.~Fang, Z.~Zhu, S.~Xu, C.~Wang, and C.~Lu, ``Rgb matters: Learning
  7-dof grasp poses on monocular rgbd images,'' in \emph{Proc. IEEE Int. Conf.
  Robot. and Automat.}, 2021, pp. 13\,459--13\,466.

\bibitem{Zhang.2021}
H.~Zhang, J.~Peeters, E.~Demeester, and K.~Kellens, ``A cnn-based grasp
  planning method for random picking of unknown objects with a vacuum
  gripper,'' \emph{J. of Intell. {\&} Robot. Syst.}, vol. 103, no.~4, pp.
  1--19, 2021.

\bibitem{C.Eppner.2020}
C.~Eppner, A.~Mousavian, and D.~Fox, ``Acronym: A large-scale grasp dataset
  based on simulation,'' in \emph{Proc. IEEE Int. Conf. Robot. and Automat.},
  2021, pp. 6222--6227.

\bibitem{Mahler.20.09.2017}
J.~Mahler, M.~Matl, X.~Liu, A.~Li, D.~Gealy, and K.~Goldberg, ``Dex-net 3.0:
  Computing robust vacuum suction grasp targets in point clouds using a new
  analytic model and deep learning,'' in \emph{Proc. IEEE Int. Conf. Robot. and
  Automat.}, 2018, pp. 5620--5627.

\bibitem{Kleeberger.12.01.2021}
K.~Kleeberger \emph{et~al.}, ``Transferring experience from simulation to the
  real world for precise pick-and-place tasks in highly cluttered scenes,'' in
  \emph{Proc. IEEE/RSJ Int. Conf. Intell. Robots and Syst.}, 2020, pp.
  9681--9688.

\bibitem{Morrison.03.03.2020}
D.~Morrison, P.~Corke, and J.~Leitner, ``Egad! an evolved grasping analysis
  dataset for diversity and reproducibility in robotic manipulation,''
  \emph{IEEE Robot. and Automat. Lett.}, vol.~5, no.~3, pp. 4368--4375, 2020.

\bibitem{M.Sundermeyer.2020}
{M. Sundermeyer}, {A. Mousavian}, {R. Triebel}, and {D. Fox},
  ``Contact-graspnet: Efficient 6-dof grasp generation in cluttered scenes,''
  in \emph{Proc. IEEE Int. Conf. Robot. and Automat.}, 2021, pp.
  13\,438--13\,444.

\bibitem{P.Hopfgarten.2020}
P.~Hopfgarten, J.~Auberle, and B.~Hein, ``Grasp area detection of unknown
  objects based on deep semantic segmentation.'' in \emph{Proc. IEEE Int. Conf.
  Automat. Sci. and Eng.}, 2020, pp. 804--809.

\bibitem{Zhao.28.02.2020}
B.~Zhao, H.~Zhang, X.~Lan, H.~Wang, Z.~Tian, and N.~Zheng, ``Regnet:
  Region-based grasp network for end-to-end grasp detection in point clouds,''
  in \emph{Proc. IEEE Int. Conf. Robot. and Automat.}, 2021, pp.
  13\,474--13\,480.

\bibitem{Eppner.11.12.2019}
C.~Eppner, M.~M. Arsalan, and D.~Fox, ``A billion ways to grasps - an
  evaluation of grasp sampling schemes on a dense, physics-based grasp data
  set,'' in \emph{Proc. Int. Symp. of Robot. Res.}, 2019, p. 890–905.

\bibitem{Gabriel.2020}
F.~Gabriel, M.~Fahning, J.~Meiners, F.~Dietrich, and K.~Dr{\"o}der, ``Modeling
  of vacuum grippers for the design of energy efficient vacuum-based handling
  processes,'' \emph{Prod. Eng.}, vol.~14, no. 5-6, pp. 545--554, 2020.

\bibitem{Provot1995}
X.~Provot, ``Deformation constraints in a mass-spring model to describe rigid
  cloth behaviour,'' in \emph{Proc. Graph. Interface Conf.}, 1995, pp.
  147--154.

\bibitem{Bernardin.2019}
A.~Bernardin, C.~Duriez, and M.~Marchal, ``An interactive physically-based
  model for active suction phenomenon simulation,'' in \emph{Proc. IEEE/RSJ
  Int. Conf. on Intell. Robots and Sys.}, 2019, pp. 1466--1471.

\bibitem{danielczuk_2019}
M.~Danielczuk \emph{et~al.}, ``Segmenting unknown 3d objects from real depth
  images using mask r-cnn trained on synthetic data,'' in \emph{Proc. IEEE Int.
  Conf. on Robot. and Automat.}, 2019, pp. 7283--7290.

\bibitem{Back.23.09.2021}
S.~Back \emph{et~al.}, ``Unseen object amodal instance segmentation via
  hierarchical occlusion modeling,'' \emph{arXiv:2109.11103}, 2022.

\bibitem{Zuo.2021}
G.~Zuo, J.~Tong, H.~Liu, W.~Chen, and J.~Li, ``Graph-based visual manipulation
  relationship reasoning network for robotic grasping,'' \emph{Frontiers in
  Neurorobotics}, vol.~15, 2021.

\bibitem{Li.2021}
T.~Li, F.~Wang, C.~Ru, Y.~Jiang, and J.~Li, ``Keypoint-based robotic grasp
  detection scheme in multi-object scenes,'' \emph{Sensors}, vol.~21, no.~6, p.
  2132, 2021.

\bibitem{Z.Hu.2021}
{Z. Hu}, {R. Hou}, {J. Niu}, {X. Yu}, {T. Ren}, and {Q. Li}, ``Object pose
  estimation for robotic grasping based on multi-view keypoint detection,'' in
  \emph{Proc. Int. Symp. on Parallel and Distrib. Process. with Appl.}, 2021,
  pp. 1295--1302.

\bibitem{Murray.op.1994}
R.~Murray, Z.~Li, and S.~Sastry, \emph{A mathematical introduction to robotic
  manipulation}.\hskip 1em plus 0.5em minus 0.4em\relax Boca Raton: CRC Press,
  1994.

\bibitem{Makoviychuk.24.08.2021}
V.~Makoviychuk \emph{et~al.}, ``Isaac gym: High performance gpu-based physics
  simulation for robot learning,'' \emph{arXiv:2108.10470}, 2021.

\bibitem{Mockus77}
J.~Mo{\v{c}}kus, ``On bayesian methods for seeking the extremum,'' in
  \emph{Proc. Optim. Techn. IFIP Tech. Conf. Novosibirsk}, 1974, pp. 400--404.

\bibitem{NVIDIADeveloper.2019}
\BIBentryALTinterwordspacing
{NVIDIA Developer}, ``Nvidia isaac sim,'' 2019. [Online]. Available:
  \url{https://developer.nvidia.com/isaac-sim}
\BIBentrySTDinterwordspacing

\bibitem{Zhou2018}
Q.~Zhou and V.~Park, J.and~Koltun, ``{Open3D}: {A} modern library for {3D} data
  processing,'' \emph{arXiv:1801.09847}, 2018.

\bibitem{maskrcnn}
K.~He, G.~Gkioxari, P.~Dollar, and R.~Girshick, ``Mask {R-CNN},'' in
  \emph{Proc. IEEE Int. Conf. on Comput. Vision}, 2017, pp. 2980--2988.

\end{thebibliography}

%\printbibliography %Prints bibliography
%\begin{thebibliography}
%\bibliographystyle{IEEEtran}
%\bibliography{refs}%,egbib}
%\end{thebibliography}

\end{document}